\documentclass{article}
\usepackage{spconf,amsmath,graphicx}
\usepackage{algorithmic,algorithm}
\usepackage[marginal]{footmisc}
\usepackage{enumitem}
\usepackage{times}
\usepackage{amsthm}
\usepackage{multirow}

\usepackage{amssymb}

\title{RESOLUTION-INVARIANT PERSON REID BASED ON FEATURE TRANSFORMATION AND SELF-WEIGHTED ATTENTION}
\name{Ziyue Zhang$^1$ \quad Shuai Jiang$^{1\star}$ \quad Congzhentao Huang$^1$ \quad Richard Yi Da Xu$^{1}$}
\address{$^1$University of Technology Sydney}
\begin{document}
%
\maketitle
%

\begin{abstract}
Person Re-identification (ReID) is a critical computer vision task which aims to match the same person in images or video sequences. Most current works focus on settings where the resolution of images is kept the same. However, the resolution is a crucial factor in person ReID, especially when the cameras are at different distances from the person or the camera's models are different from each other. In this paper, we propose a novel two-stream network with a lightweight resolution association ReID feature transformation (RAFT) module and a self-weighted attention (SWA) ReID module to evaluate features under different resolutions. RAFT transforms the low resolution features to corresponding high resolution features. SWA evaluates both features to get weight factors for the person ReID. Both modules are jointly trained to get a resolution-invariant representation. Extensive experiments on five benchmark datasets show the effectiveness of our method. For instance, we achieve Rank-1 accuracy of 43.3\% and 83.2\% on CAVIAR and MLR-CUHK03, outperforming the state-of-the-art.
\end{abstract}


\section{Introduction}
\label{sec:intro}
Person ReID is a vital computer vision task that aims to recognize the same person across the images or video sequences taken by different cameras \cite{zheng2016person, ye2020deep}.
Due to the success of convolutional neural networks (CNNs), a variety of learning-based methods \cite{chang2018multi, sun2018beyond, zheng2017unlabeled} have been proposed to address person ReID problem and achieved good performance on many benchmarks.
Nevertheless, with the presence of people occlusions \cite{huang2018adversarially,hou2019vrstc}, viewpoint changes \cite{zheng2017unlabeled,zheng2019joint}, pose changes \cite{zhu2019progressive,qian2018pose}, and even illumination changes \cite{zeng2019illumination,9190796}, person reID remains a very challenging task.

However, these methods are mostly developed under the settings where both query and gallery images are of high resolution (HR), which is not always true in practice since people may be captured by a camera in one resolution and then reappear under another camera in a drastically different resolution. In practical settings, query images are usually of low resolution (LR) captured by a surveillance system, while gallery images are mostly of HR which are chosen carefully. 
In practice, resolution changes can significantly affect the ReID results because of less discriminative appearance information of LR images and the gap between image features from different resolutions. Therefore, it is a necessity to study algorithms for handling resolution changes with high robustness in Person ReID.

To address the above problem, existing methods \cite{jiao2018deep, chen2019learning, mao2019resolution,li2019recover} either use super-resolution (SR) models or generative adversarial network (GAN) to convert LR images into their corresponding HR images, followed by a person ReID model. However, these two kinds of methods suffer from different problems. SR is a one-on-one scheme, which means it is unable to handle arbitrary inputting resolution. Multiple SRs may be trained to deal with multiple inputting/outputting resolutions. However, the number of resolutions is still finite and needs to be pre-defined.
GAN based methods require a large number of resources and training time, making them impractical in many resource-limited scenarios. Both of them regard generating HR images and person ReID as two separately consecutive phases, ignoring the relations between the features in both phases.

To better resolve the above challenges, in this paper, we propose a novel network with the following three highlights:
\begin{enumerate}[topsep = 5 pt]
\setlength{\itemsep}{4pt}
\setlength{\parsep}{0pt}
\setlength{\parskip}{0pt}
\item
The input of the proposed network is two-stream, one for HR images and the other for LR images.
\item
We design a module to transform LR image features to their corresponding HR image features, rather than the transformation between images themselves. The module manages to connect the relations between ReID features from different resolutions.
\item
A self-weighted attention ReID module is proposed to evaluate the quality of each input HR and LR ReID features, making it capable of merging HR and LR features with a weighted distance metric.
\end{enumerate}


\begin{figure*}[tbp]
    \centering
    \includegraphics[width = 1\textwidth]{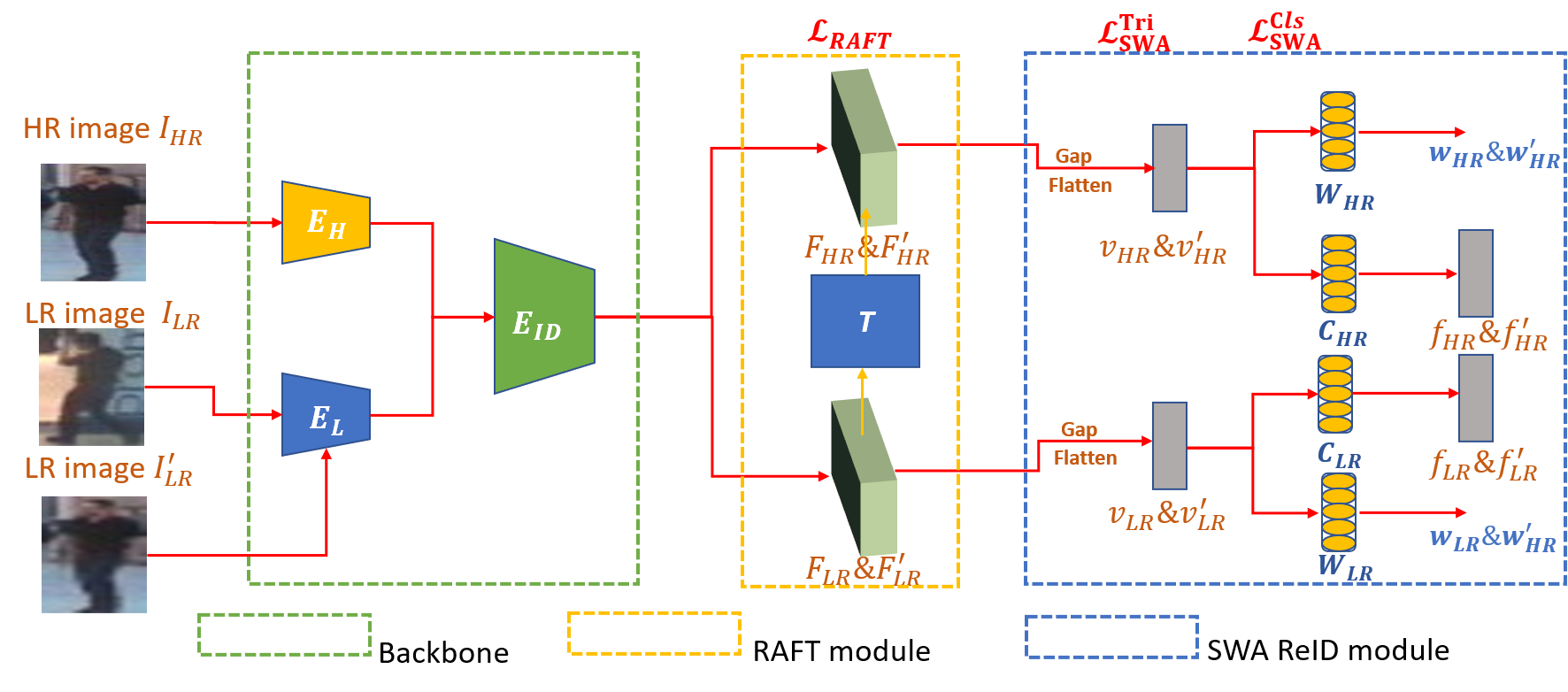}
    \caption{The architecture overview of our FTWA Network with the inputs and outputs of it. It contains (1) backbone, (2) RAFT module and (3) SWA ReID module.}
    \label{fig:wholeModels}
\end{figure*}

\section{Method}\label{sec:Method}


We proposed a two-stream input and two-stream output network structure. 
Fig \ref{fig:wholeModels} shows the overall model structure of our FTWA (Feature Transformation and self-weighted Attention) model. It includes three main parts: (1) two-stream input backbone, (2) resolution association ReID feature transformation (RAFT) module and (3) self-weighted attention (SWA) ReID module.

\subsection{Backbone}

The backbone model consists of two parts. The first part includes LR image encoder ($E_{L}$) and HR image encoder ($E_{H}$), which is for shallow image features extraction. These two encoders have the same structure but independent parameters. The second part is the ReID encoder ($E_{ID}$), which is to get the ReID features for matching. The whole backbone model is based on a Resnet50 \cite{he2016deep}. The first part of the backbone consists of the first convolutional layer, followed by the first block of Resnet50. The second part is the last three blocks of Resnet50.

The backbone takes two inputs. One of them is HR images ($I_{HR}$), and the other is LR images ($I_{LR}$). 
\begin{equation}
\label{backbone}
\begin{aligned}
F_{HR} = E_{ID}(E_{H}(I_{HR})),\\
F_{LR} = E_{ID}(E_{L}(I_{LR})),\\
F^{'}_{LR} = E_{ID}(E_{L}(I^{'}_{LR}))
\end{aligned}
\end{equation}
For each HR input image, we use down-sample methods to generate the LR version ($I^{'}_{LR}$) of it. 
As shown in equation \ref{backbone}, we then send both of them to the backbone model to get the HR and corresponding LR ReID features ($F_{HR}$ and $F{'}_{LR}$). 
For each LR input image, we send itself to the backbone model to get the LR ReID feature ($F_{LR}$).

\subsection{Resolution association ReID feature transformation module}

To address the input images' resolution variation, we construct a resolution association ReID feature transformation (RAFT) module $T$ to transform LR image features to corresponding HR image features. Inspired by the lightweight structure in \cite{Hui-IMDN-2019}, we construct RAFT in a similar way. However, unlike traditional GAN or super-resolution based resolution ReID methods using LR images as input, our RAFT module takes LR image features as input generated by the backbone's ReID encoder $E_{ID}$. 

\begin{figure}[tbp]
    \centering
     \includegraphics[width = 0.4\textwidth]{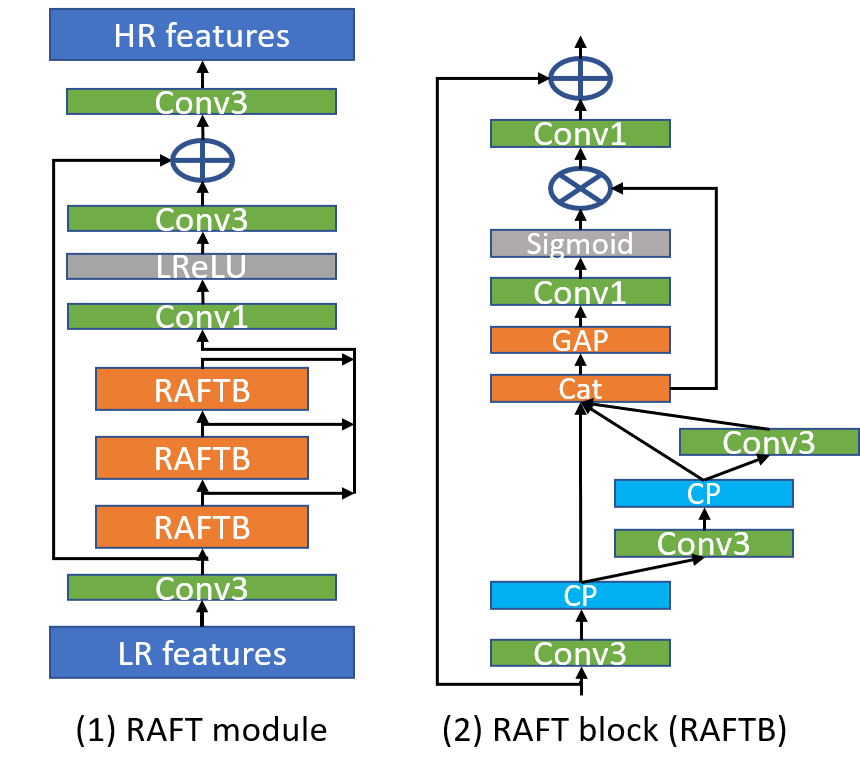}
    \caption{The structure of our proposed RAFT module and RAFT block (RAFTB).}
    \label{fig:RAFT}
\end{figure}

As shown and describe in Fig \ref{fig:RAFT}, we construct the RAFT module in this specific way. The left of the picture is the RAFT module, which consists of several convolutional layers and RAFT blocks (RAFTB). The right part of the picture is RAFTB, which consists of multiple convolutional layers. The Conv means convolutional layer and the number followed by is kernel size of this layer. The LReLU and Sigmoid are Leaky ReLU and Sigmoid activation layer. The CAT means Channel Concatenating layer and GAP means Global Average Pooling layer. The CP means channel split operation, which split the input tensors into two tensors on average by channel. The plus and multiplication sign in the picture means element-wise add and multiply operation. Each convolutional layer except the Conv1 layer in RAFTB is followed by a leaky ReLU layer, which we do not show in the picture for concise.

By this design, we can generate HR image features using few parameters, which is more resource-saving. To train the RAFT module, we need to send fake LR image features ($F^{'}_{LR}$) into it to get the corresponding HR version. The RAFT aims to make the generated features as similar as possible to original HR image features ($F_{HR}$). So we can get the loss for RAFT as follows:
\begin{equation}
\label{RAFTLoss}
\mathcal{L}_{RAFT} = \lVert F_{HR} - T(F^{'}_{LR})\rVert_{1},
\end{equation}
where we use L1 norm as our loss function.
We also need to send original LR image features ($F_{LR}$) to RAFT module to get fake HR image features ($F^{'}_{HR}$) as follows:
\begin{equation}
\label{fakeHR}
F^{'}_{HR} = T(F_{LR}).
\end{equation}
Then we can use these four image features ($F_{HR}$, $F_{LR}$, $F^{'}_{HR}$, $F^{'}_{LR}$) to train the SWA ReID module.

\subsection{Self-weighted attention ReID module}

To use the relation between different resolution image features, we design a self-weighted attention (SWA) ReID module. As shown in fig \ref{fig:wholeModels}, we adopt a Global Average Pooling (GAP) layer on each resolution image ReID features and then flatten them to get corresponding the person ReID latent vector ($v_{HR}$,$v_{LR}$,$v{'}_{HR}$,$v{'}_{LR}$). Finally, for both HR and LR stream, we construct a quality evaluator ($W_{HR}$,$W_{LR}$) and an identity classifier ($C_{HR}$,$C_{LR}$) separately. We then send HR and LR vectors to the corresponding evaluator and classifier to get vector quality weight and identity factor as follows:

\begin{equation}
\label{weightHR}
\begin{aligned}
w_{HR} = W_{HR}(v_{HR}), w^{'}_{HR} = W_{HR}(v^{'}_{HR}),\\
w_{LR} = W_{LR}(v_{LR}), w^{'}_{LR} = W_{LR}(v^{'}_{LR}),\\
f_{HR} = C_{HR}(v_{HR}), f^{'}_{HR} = C_{HR}(v^{'}_{HR}),\\
f_{LR} = C_{LR}(v_{LR}), F^{'}_{LR} = C_{LR}(v^{'}_{LR}).
\end{aligned}
\end{equation}

To classify the person's identity with different resolutions, we adopt two loss functions. The first one is the SWA identity classification loss $\mathcal{L}_{\text{SWA}}^{\text{Cls}}$. We use softmax cross-entropy loss ($L^{Cls}$) between the classification prediction and the corresponding ground-truth one hot vector as traditional classification loss. The SWA version we design based on $L^{Cls}$ is as follows:
\begin{equation}
\label{SWAClsLoss}
\begin{aligned}
\mathcal{L}^{\text{Cls}}_{SWA} = &\frac{w_{HR}*L^{Cls}(f_{HR}) + w^{'}_{LR}*L^{Cls}(f^{'}_{LR})}{w_{HR}+w^{'}_{LR}} \\
& + \frac{w_{LR}*L^{Cls}(f_{LR}) + w^{'}_{HR}*L^{Cls}(f^{'}_{HR})}{w_{LR}+w^{'}_{HR}}.
\end{aligned}
\end{equation}
In this loss function, the model will be optimized by each input pair of image features and its corresponding synthetic one. 

The second one is the SWA triplet loss $\mathcal{L}_{\text{SWA}}^{\text{Tri}}$ for person ReID vector similarity learning. 
The traditional triplet loss function $\mathcal{L}^{\text{Tri}}$, which makes the distance of vectors closer between the same identity and further apart between different identities, can be defined as follows: 
\begin{equation}
\label{l-reid-tri}
\mathcal{L}^{\text{Tri}}(v) = \sum_{v_{a}, v_{p}, v_{n}\in v \atop y_{a}=y_{p} \neq y_{n}}\left[m+D_{a, p}-D_{a, n}\right]_{+},
\end{equation}
where $y_{a}, y_{p}, y_{n}$ is the corresponding truth ID of sample $v_{a},v_{p},v_{n}$. $D_{a, p}$ and $D_{a, n}$ are the Euclidean distance between the anchor vector $v_a$ and positive sample vector $v_p$ (same identity sample) and that between the anchor vector $v_a$ and negative sample vector $v_n$ (different identity sample), $m$ is a margin parameter constraining the maximal distance between the anchor and negative samples and $[x]_{+} = \max(0, x)$.
Based on the above loss, we construct our proposed SWA triplet loss as follows:
\begin{equation}
\label{SWATriLoss}
\begin{aligned}
&\mathcal{L}^{\text{Tri}}_{SWA} = \frac{\mathcal{L}^{\text{Tri}}_{HR}+\mathcal{L}^{\text{Tri}}_{LR}}{w_{HR}*w^{'}_{HR}+w_{LR}*w^{'}_{LR}},\\
&\mathcal{L}^{\text{Tri}}_{HR} = w_{HR}*w^{'}_{HR}*L^{Tri}(v_{HR}\cup v^{'}_{HR}),\\
&\mathcal{L}^{\text{Tri}}_{LR} = w_{LR}*w^{'}_{LR}*L^{Tri}(v_{LR}\cup v^{'}_{LR}).\\
\end{aligned}
\end{equation}
By using SWA version loss, the model will be optimized by the weighted triplet loss of every HR image feature pairs and LR image feature pairs.
Hence the full loss function for training our FTWA Network is as follows:
\begin{equation}
\label{lossWhole}
\mathcal{L}_{\text{ReID}} =\lambda_{1} \mathcal{L}_{\text{SWA}}^{\text{Cls}}+\lambda_{2} \mathcal{L}_{\text{SWA}}^{\text{Tri}}+\lambda_{3} \mathcal{L}_{RAFT},
\end{equation}
where $\lambda_{1}$, $\lambda_{2}$ and $\lambda_{3}$ are the weights allocated for the above three loss functions. 

\section{Experiments}\label{sec:Experiment}
\subsection{Dataset and Evaluation Protocol}
We evaluate our method on four datasets, which are CUHK03 \cite{li2014deepreid}, CAVIAR \cite{cheng2011custom}, VIPeR \cite{gray2008viewpoint} and Market-1501 \cite{zheng2015scalable}. In these datasets, CAVIAR is composed of two different resolution images captured by two cameras. However, CUHK03, VIPeR and Market-1501 only include one resolution images. So following SING \cite{jiao2018deep}, we construct the synthetic multiple resolution (MLR) datasets (MLR-CUHK03, MLR-VIPeR, MLR-Market-1501). We down-sample images taken by one camera by a randomly selected down-sampling rate $r\in{2, 3, 4}$, while the images taken by the other cameras remain unchanged. 
We evaluate our method by the setting \cite{jiao2018deep} where the query set is composed of LR images while the gallery set contains HR images. In our experiments, the standard single-shot person ReID setting and the average cumulative match characteristic (CMC) Rank metric is adopted.

\subsection{Implementation Details}
We implement our model with Pytorch. We adopt the ResNet-50 \cite{he2016deep} pre-trained on ImageNet as our backbone. By modifying the last layer stride to be 1 in the backbone Resnet50, the model can make final output features have more abundant information.
The batch size is set as 64. We use Adam optimizer and set both the weight decay factor and weight decay bias factor as 0.0005. The base learning rate is 0.0007, with a linear learning rate scheduler. The total training epoch number is 120. For the epoch step in [40, 70], we decrease the learning rate by a decay factor equals to 0.2.
For some hidden parameters, we set weight parameters $\lambda_{1}$ as 3.0, $\lambda_{2}$ as 1.0 and $\lambda_{3}$ as 0.1. We set the margin value $m$ in the triplet loss as 0.3.   

\subsection{Comparison with state-of-the-art methods}
We exploit the MLR-CUHK03 (MLR-C), CAVIAR, MLR-VIPeR (MLR-V) and MLR-Market-1501 (MLR-M) to evaluate the accuracy of our model compared with other current state-of-the-art MLR person ReID deep learning methods. Experimental results are presented in Table \ref{table1}.

\begin{table}[!h]
\footnotesize
\caption{Comparison with state-of-the-art MLR ReID methods.}
\label{table1}
\vspace*{2mm}
\centering
\resizebox{0.5\textwidth}{!}{
\begin{tabular}{|l|l|l|l|l|l|l|l|l|l|}
\hline
Model  & \multicolumn{2}{l|}{MLR-C} & \multicolumn{2}{l|}{CAVIAR} & \multicolumn{2}{l|}{MLR-V} & \multicolumn{2}{l|}{MLR-M} \\ \cline{2-9} 
 & R1 & R5 & R1 & R5 & R1 & R5 & R1 & R5 \\ \hline
SING \cite{jiao2018deep}  & 67.7 & 90.7 & 33.5 & 72.7 & 33.5 & 57.0 & 74.4 & 87.8 \\ \hline
CRS-GAN \cite{wang2018cascaded}  & 71.3 & 92.1 & 34.7 &  72.5 & 37.2 &  62.3 &76.4 & 88.5 \\ \hline
RIFE \cite{mao2019resolution}  & 73.3 & 92.6 & 36.4 & 72.0 & 41.6 & 64.9 & 66.9 & 84.7 \\ \hline
RAIN \cite{chen2019learning}  & 78.9 & 97.3 & 42.0 & \textbf{77.3} & 42.5 & 68.3 & - & - \\ \hline
CRGAN \cite{li2019recover}  & 82.1 & 97.4 & 42.8 & 76.2 & 43.1 & 68.2 & 83.7 & 92.7 \\ \hline
Ours & \textbf{83.2}  & \textbf{97.8} & \textbf{43.3} & 77.1 & \textbf{43.2} & \textbf{68.5} & \textbf{84.4} & \textbf{93.7} \\ \hline
\end{tabular}
}
\end{table}
As shown in the table, our proposed method is the best overall all methods on these four datasets. Take MLR-CUHK03 as an example, and it achieves at most 83.2\% Rank1 accuracy, which outperforms the best competitors \cite{chen2019learning, li2019recover} by 1\% to 4\%. The performance gains can be ascribed as follows: First, unlike most existing MLR person ReID methods, our model transforms the LR features into corresponding HR features rather than transform LR image into corresponding HR image. Second, our method performs MLR person ReID end to end. Third, our model use both original and synthetic image features for ReID with self-weighted attention. This experiment demonstrates that our model is superior to state-of-the-art methods with respect to ReID accuracy.

\subsection{Ablation study}

To illustrate the effectiveness of the modules proposed in our work, we design four variant settings of our model. The baseline model is a Resnet50 with ReID classification loss and triplet loss. The $\text{FTWA}_{B}$ is only the two-stream input backbone model without RAFT and SWA ReID loss. The $\text{FTWA}_{R}$ uses RAFT to transform image features but does not use SWA ReID loss. The $\text{FTWA}$ is our whole FTWA model with RAFT and SWA ReID loss. The results of the baseline and these variants on MLR-CUHK03 are shown in Table \ref{table2}.
\begin{table}[h]\tiny
\caption{Performance of baseline and variants of proposed model on MLR-CUHK03 dataset.}
\label{table2}
\vspace*{2mm}
\centering
\resizebox{0.25\textwidth}{11mm}{%
\begin{tabular}{|l|l|l|}
\hline
Model  & \multicolumn{2}{l|}{MLR-CUHK03} \\ \cline{2-3} 
 & R1 & R5 \\ \hline
Baseline & 46.3 & 77.2  \\ \hline
$\text{FTWA}_{B}$ & 52.6  & 78.7  \\ \hline
$\text{FTWA}_{R}$ & 82.3  & 97.4  \\ \hline
$\text{FTWA}$ & \textbf{83.2}  & \textbf{97.8}  \\ \hline
\end{tabular}%
}
\end{table}

As shown in the table, the baseline model can only get 46.3\% Rank1 accuracy on MLR-CUHK03. This is because of no process on the domain gap between different resolutions. $\text{FTWA}_{B}$ can improve a little accuracy since the backbone model process the HR and LR image input separately. $\text{FTWA}_{R}$ improve so much accuracy than $\text{FTWA}_{B}$, which is because the model constructs the relation between HR image features and LR image features through RAFT. Finally, the whole $\text{FTWA}$ can get the highest accuracy overall all settings. The reason is that the model can evaluate the quality of the person ReID vector by using SWA ReID loss. The model can then better combine the HR and LR information of each image input (no matter HR or LR image input) together.

\section{Conclusion}
This paper has proposed a novel FTWA model to learn the common representation features for cross-resolution person images. In our method, the FTWA model consists of three main components, which are (1) two input backbone for processing HR and LR input independently, (2) a lightweight and effective resolution association ReID feature transformation module and (3) the self-weighted attention ReID module. We transform the LR person image features into a corresponding HR version in our method and use a novel weighted loss to merge them. Comprehensive experiments on the challenging cross-resolution person ReID datasets have demonstrated that our approach outperforms the state-of-the-art methods.

\bibliographystyle{IEEEbib}
\bibliography{refs}

\end{document}